\documentclass{article}
\usepackage[utf8]{inputenc}
\usepackage{titling}
\usepackage{lipsum}
\usepackage[space]{grffile}
\usepackage{graphicx}
\usepackage{url}
\usepackage{amsmath}
\usepackage{amssymb}
\usepackage{url}
\usepackage{float}
\usepackage[hang,flushmargin]{footmisc} 

\newcommand{\figwid}{\linewidth}

\newcommand{\Cm}{\mbox{${\bf C}$}}
\newcommand{\muv}{\mbox{$\vec{\mu}$}}

\makeatletter
\def\@setthanks{\vspace{-\baselineskip}\def\thanks##1{\@par##1\@addpunct.}\thankses}
\makeatother

\title{\Huge\bf Covapixels}
\author{{\Large Jeffrey Uhlmann}\thanks{Prepared for {\em Lucharama}-2020 (c.f., \cite{Lucharama}).}\\
Dept.\ of Electrical Engineering and Computer Science\\
University of Missouri - Columbia}
\date{}

\begin{document}

\maketitle
\vspace*{-0.25in}
\begin{abstract}
We propose and discuss the summarization of superpixel-type image tiles/patches
using mean and covariance information. We refer to the resulting objects
as {\em covapixels}. 
\end{abstract}


\section{Introduction}

Various approaches have been examined for reducing
the input complexity of data to be processed by 
multilayer learning architectures. This is particularly
of interest for the processing of large, high-resolution 
images by neural-inspired networks. Specifically, there
is need to reduce the effective pixel-complexity of large
high-definition and ultra-high definition images to 
permit practical training on large image datasets.

The simplest approach for reducing pixel complexity
is to perform simple tile-based decimation, e.g., 
perform bicubic scaling to a lower resolution.
A more sophisticated approach involves performing
a non-regular decimation into non-rectangular tiles,
referred to as {\em superpixels} \cite{RenMalik},
that are constrained to conform to salient structures
of the image. Figures 1 and 2 provide representative
examples\footnote{Superpixel tessellations depicted in figures
were computed using the online segmentation tool 
of \cite{slic}.}. The efficacy of superpixel methods
has been demonstrated in a variety of applications,
but various limitations have also been identified \cite{yang}.

\begin{figure}
\begin{center}
\includegraphics[width=\figwid,keepaspectratio]{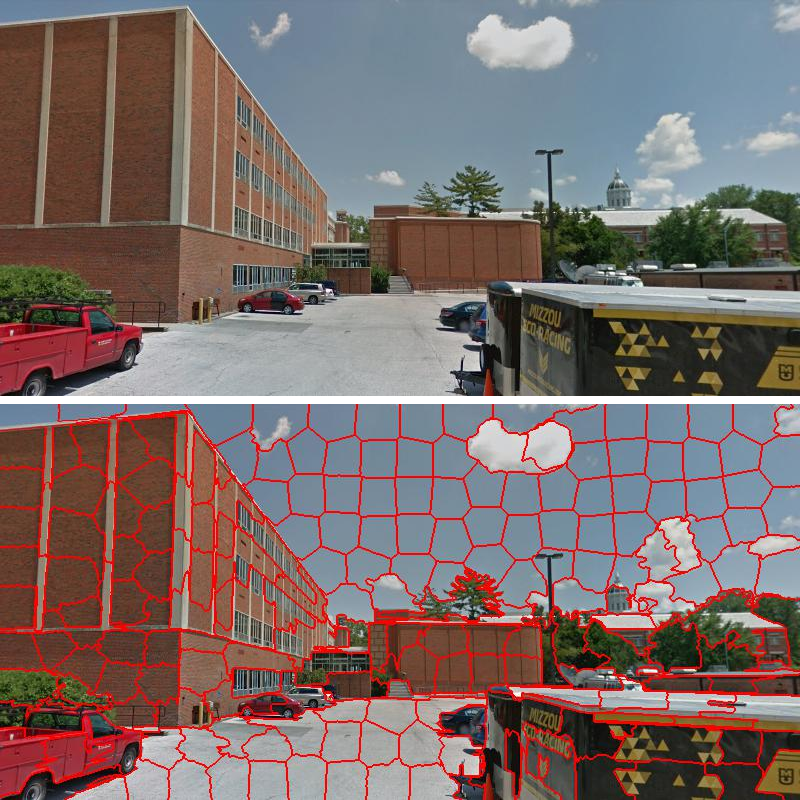}
\caption{\footnotesize Example of an image of an exterior location ({\em top}) 
and its decomposition into superpixels ({\em bottom}). Note the arbitrary
superpixel boundaries in relatively uniform regions of the sky and pavement. }
\label{fig:p-0}
\end{center}
\end{figure}

\begin{figure}
\begin{center}
\includegraphics[width=\figwid,keepaspectratio]{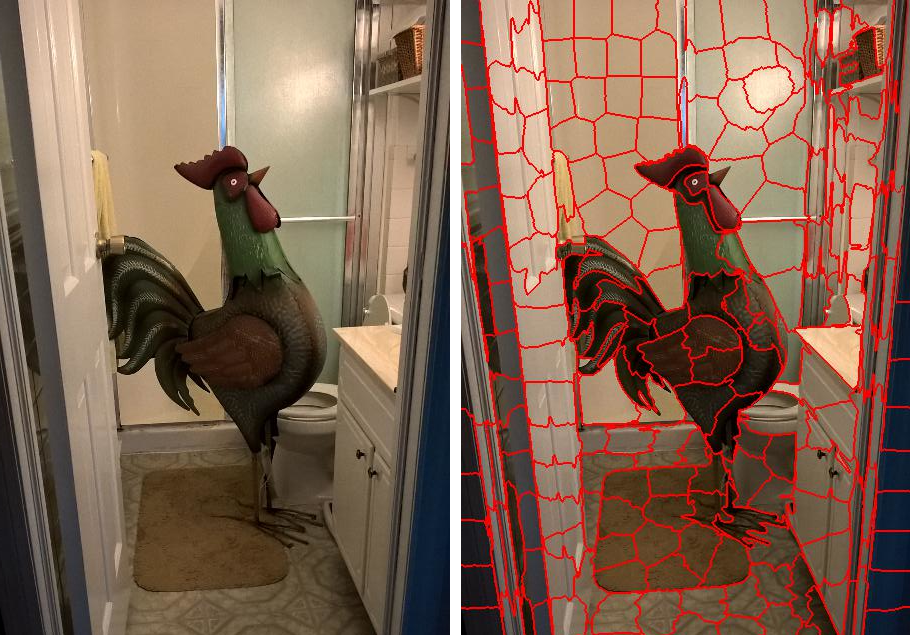}
\caption{\footnotesize Example of an image of an interior location ({\em left}) 
and its decomposition into superpixels ({\em right}). Note the arbitrary
superpixel boundaries in relatively uniform regions of the shower and
rug.}
\end{center}
\end{figure}

In this paper we briefly introduce an alternative approach, 
based on the notion of {\em covapixels} for the compressed 
represention of images and other forms of large structured 
pieces of information. In the following section we define an
example of a covapixel representation in which the analog
of a superpixel may take the form of a pair $(\muv,\Cm)$, where
$\muv$ is a $k \times 1$ vector and $\Cm$ is $k\times k$ 
symmetric (hermitian) nonnegative-definite 
matrix\footnote{This matrix may generally be represented
in a variety of simplified or compressed forms, e.g., as the triangular 
Cholesky square root.}.

\section{Covapixels}

The motivation underpinning use of superpixels is the desire
to obtain the data-reduction advantages of tile-based 
resolution reduction while minimizing the potential loss of
important detail information. Unfortunately, the irregular
boundaries of superpixels can have the effect of introducing
spurious detail information. More specifically, the imposition
of an artificial size or area constraint on superpixels will
tend to introduce relatively complex artificial segmentation 
boundaries within large homogeneous areas of a given
image. In other words, spurious feature entropy is introduced 
into the covapixel decomposition of the image.

Upon further reflection, it becomes clear that whatever
generalization of a pixel is defined must somehow 
produce a relatively homogeneous spatial tessellation in which 
each generalized pixel encodes a region that is as locally
uniform as possible in terms of image detail. The problem
that arises with superpixels is the representational complexity
of their boundaries, which then increases the input complexity 
of whatever system / network is expected to manipulate and
process them.

We proprose to address many of the limitations of previous 
methods for image complexity reduction by decomposing images 
in a way that encodes local image information in a simpler form
that admits use of standard operators used in tracking and 
control applications. Specifically, we represent a covapixel
as a vector and matrix pair that summarizes
a given tile/region of an image in which, for example, 
the vector defines the location (and possibly other attributes)
of the region while the matrix encodes some measure of the 
spatial extent of the region and/or the distribution associated 
with some measure of the content (feature attributes) of the region.  

The reason for adopting a {\em mean} and {\em covariance} 
representation is to permit the processing system/network to 
use data fusion operators such as the Kalman filter (KF) 
update (and its inverse/information form) \cite{kf}; 
Covariance Intersection (CI); Covariance Union (CU); 
Covariance Addition (CA); and their variants\footnote{See
the appendices of \cite{jkucu} for a unified discussion of
CI, CU, and CA.} (\cite{jkucu,gencu}) 
for most or all internal processing of covapixel information.
More intuitively, the complex boundary of a superpixel is replaced
with a mean and covariance statistical representation that can be
interpreted (though necessarily so) as a Gaussian probability 
distribution, e.g., as informally depicted in Figure 3.

\begin{figure}
\begin{center}
\includegraphics[width=\figwid,keepaspectratio]{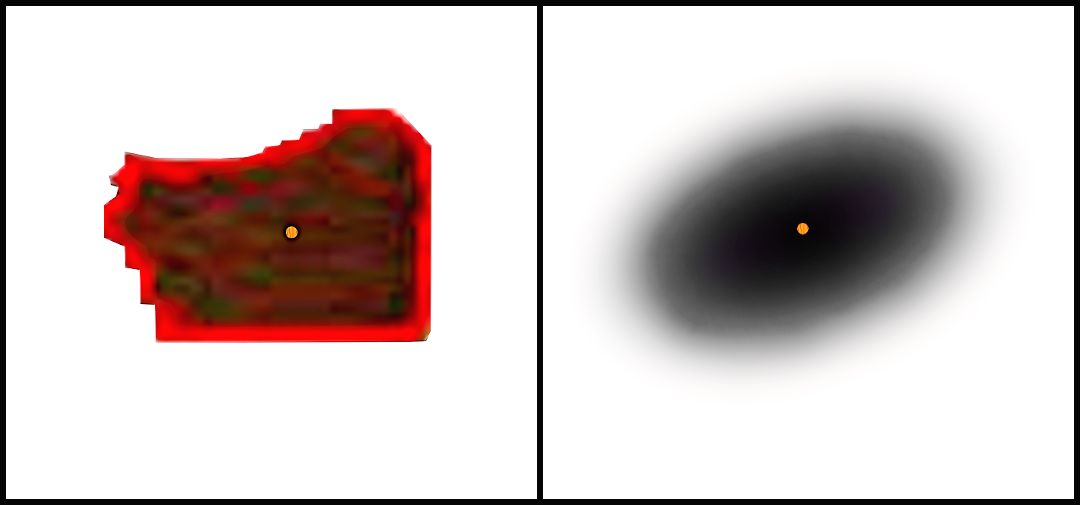}
\caption{\footnotesize Superpixel ({\em left}) from a red-brick wall in Figure 1 and
its summarizing covapixel ({\em right}). The gold circles correspond respectively ({\em l-r})
to the centroid of the superpixel and the mean vector $\muv$ of its 
covapixel representation.}
\end{center}
\end{figure}

The key feature of the mean-covariance covapixel representation is
that its information is in a form that can be directly used by standard
data fusion operators as depicted in Figure 4. 

\begin{figure}
\begin{center}
\includegraphics[width=\figwid,keepaspectratio]{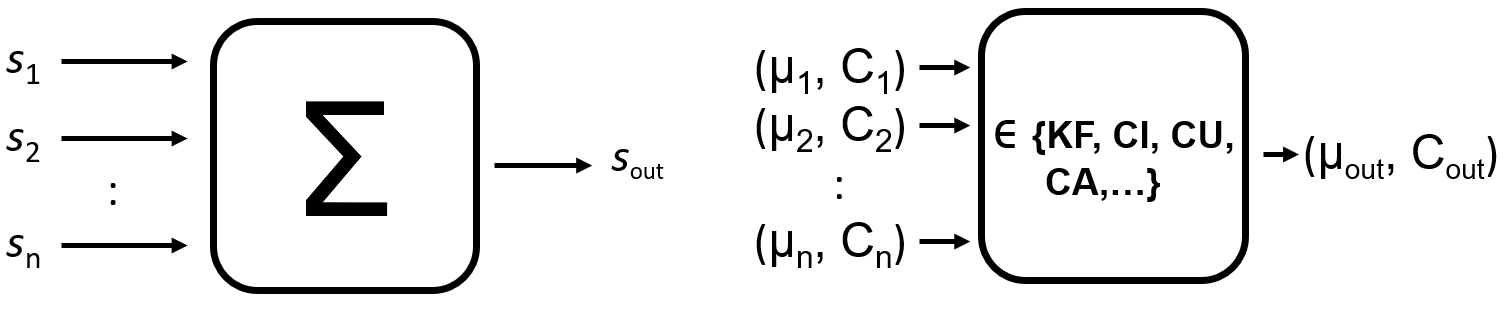}
\caption{\footnotesize Processing of scalars derived from a superpixel ({\em left}) 
versus processing of mean and covariance pairs using conventional data fusion
operators ({\em right}). This generalization can potentially be applied
more generally to the formulation of neurons in artificial neural network-type
architectures.}
\end{center}
\end{figure}

\section{Discussion}

We have briefly introduced and discussed a generalization of pixel
information that can be expressed in the mean and covariance form
widely used in filtering and control applications. This generalization is
referred to as a {\em covapixel} because it can encode arbitrary
state information in terms of a state vector and an associated error
covariance matrix\footnote{It should be noted that the term 
{\em covariance matrix} does not necessarily denote the second
central moment of a probability distribution. Specifically, it may represent
an upper bound on that moment \cite{jkucu} or could be interpreted to define an
elliptical/ellipsoidal bounded region/volume \cite{gencu2}.}. 

Although we have focused
on applications to image processing, the replacing of scalars with mean
and covariance pairs could even be applied to the elements of matrices and
tensors, e.g., to maintain covariance estimates of numerical error that 
accrues from the application of linear algebra operators. This of course
suggests the potential use of units of this kind in neural network-type
architectures which would intrinsically process mean and covariance pairs rather
than scalars.  

One of the key potential benefits of covapixels is their simpler parametric
representation relative to the complex boundaries of superpixels. It can be
anticipated that this simpler representation will reduce the amount of spurious 
implicit feature detail that can result from complex boundaries. The critical
question that remains to be answered is whether the loss of precise 
boundary detail results in a loss of salient information that undermines the 
practical utility the approach\footnote{If this is the case, then an alternative in the 
opposite direction would be to represent superpixel information in the form
an adjacency matrix representing a graph that may include non-rigid joints,
which would incur a significant increase in complexity but would admit various
tools from linear algebra \cite{mm} to be applied for the dynamic 
maintenance of tessellation structures, e.g., in video applications.}.

\bibliographystyle{plain}

\end{document}